%% file: space.tex
\documentclass[sigconf]{acmart}
\usepackage{amsmath}
\usepackage{comment}
\usepackage{enumitem}
\usepackage{makecell}
\usepackage{pifont}
\newcommand{\zjh}[1]{{\color{black}#1}}

\copyrightyear{2026}
\acmYear{2026}
\setcopyright{cc}
\setcctype{by}
\acmConference[MM '26]{Proceedings of the 34th ACM International Conference on Multimedia}{November 10--14, 2026}{Rio de Janeiro, Brazil.}
\acmBooktitle{Proceedings of the 34th ACM International Conference on Multimedia (MM '26), November 10--14, 2026, Rio de Janeiro, Brazil}
\acmISBN{979-8-4007-2213-4/2026/11}
\acmDOI{10.1145/3767308.3835525}
\makeatletter
\def\@affiliationfont{\normalsize}
\makeatother

\AtBeginDocument{%
  }

\begin{document}

\title[Human-Centric Captioning]{Human-Centric Image Captioning with Subject-Centered Spatial Understanding}

\author{Bozhou Li}
\authornote{Equal contribution.}
\authornote{Work done during an internship at the Kling Team.}
\affiliation{%
  \institution{Peking University}
  \city{Beijing}
  \country{China}}
\email{2301213084@pku.edu.cn}

\author{Jiahang Zhang}
\authornotemark[1]
\authornotemark[2]
\affiliation{%
  \institution{Peking University}
  \city{Beijing}
  \country{China}}
\email{1900012780@pku.edu.cn}

\author{Yue Ding}
\affiliation{%
  \institution{Chinese Academy of Sciences}
  \city{Beijing}
  \country{China}}
\email{yueding1011@gmail.com}

\author{Yushuo Guan}
\authornote{Project leader.}
\affiliation{%
  \institution{Kling Team}
  \city{Beijing}
  \country{China}}
\email{yushuo.guan@gmail.com}

\author{Bohan Zeng}
\affiliation{%
  \institution{Peking University}
  \city{Beijing}
  \country{China}}
\email{bhzeng25@stu.pku.edu.cn}

\author{Yiyan Ji}
\affiliation{%
  \institution{Nanjing University}
  \city{Nanjing}
  \country{China}}
\email{jiyiiiyyy@gmail.com}

\author{Xinlong Chen}
\affiliation{%
  \institution{Chinese Academy of Sciences}
  \city{Beijing}
  \country{China}}
\email{chenxinlong2025@ia.ac.cn}

\author{Yang Shi}
\affiliation{%
  \institution{Peking University}
  \city{Beijing}
  \country{China}}
\email{frankyang1517@gmail.com}

\author{Yifan Dai}
\affiliation{%
  \institution{Shanghai Jiao Tong University}
  \city{Shanghai}
  \country{China}}
\email{yfandaii@sjtu.edu.cn}

\author{Yuran Wang}
\affiliation{%
  \institution{Peking University}
  \city{Beijing}
  \country{China}}
\email{yuranwang25@stu.pku.edu.cn}

\author{Chengzhuo Tong}
\affiliation{%
  \institution{Peking University}
  \city{Beijing}
  \country{China}}
\email{22009200626@stu.xidian.edu.cn}

\author{Pengfei Wan}
\affiliation{%
  \institution{Kling Team}
  \city{Beijing}
  \country{China}}
\email{wpf1987@outlook.com}

\author{Yuanxing Zhang}
\authornote{Corresponding author.}
\affiliation{%
  \institution{Kling Team}
  \city{Beijing}
  \country{China}}
\email{longo11070001@gmail.com}

\author{Wentao Zhang}
\affiliation{%
  \institution{Peking University}
  \city{Beijing}
  \country{China}}
\email{wentao.zhang@pku.edu.cn}

\renewcommand{\shortauthors}{Bozhou Li et al.}

\begin{abstract}
While multimodal large language models (MLLMs) achieve remarkable performance on generic image captioning, they frequently suffer from structural hallucinations in human-centric scenarios. 
Accurately modeling human subjects is foundational for critical downstream applications, such as accurate avatar/video/image generation and fine-grained human action understanding. 
However, these tasks require highly precise subject-centered spatial grounding, such as distinguishing egocentric left/right laterality and maintaining correct anatomical-object bindings. 
Although catastrophic for structural integrity, these localized spatial inversions are often overshadowed by overall descriptive metrics in existing benchmarks. 
To systematically expose and quantify this bottleneck, we introduce \textit{\textbf{SPACE}} (\textbf{S}ubject-centric \textbf{P}oses, \textbf{A}ppearance, and \textbf{C}haracteristics \textbf{E}valuation), a benchmark designed to evaluate \textit{subject-centered} spatial understanding. 
On SPACE, we reveal that despite strong generic perception, current MLLMs consistently fail to ground descriptions in the subject's intrinsic frame of reference. To bridge this gap, we propose a specialized data construction and alignment pipeline. 
We first extract structured spatial hints from fine-grained body-part localization to guide a two-stage caption rewriting process, yielding highly spatially-faithful training data. Furthermore, we design a rubric-based reward for Group Relative Policy Optimization (GRPO) that explicitly penalizes structurally critical spatial errors during alignment. Extensive experiments on SPACE demonstrate our framework significantly improves human-centric caption quality, particularly in subject-centered spatial reasoning, achieving performance competitive with strong closed-source models. 
Our benchmark and code are available at \url{https://github.com/JHang2020/SPACE-Eval}.

\end{abstract}


\maketitle
\input{figures/fig-benchmark-casestudy}
\input{section/0_intro}
\input{section/1_related_work}
\input{section/2_benchmark}
\input{section/3_method}

\input{section/4_exp}

\input{section/5_conclusion}

\clearpage
\begin{acks}
This work was supported by the Beijing Major Science and Technology Project (No.~Z251100008425023).
\end{acks}

\bibliographystyle{ACM-Reference-Format}
\bibliography{sample-base}

\end{document}

%% file: figures/fig-benchmark-casestudy.tex
\begin{figure*}[!ht]
    \centering
    \includegraphics[width=0.86\textwidth]{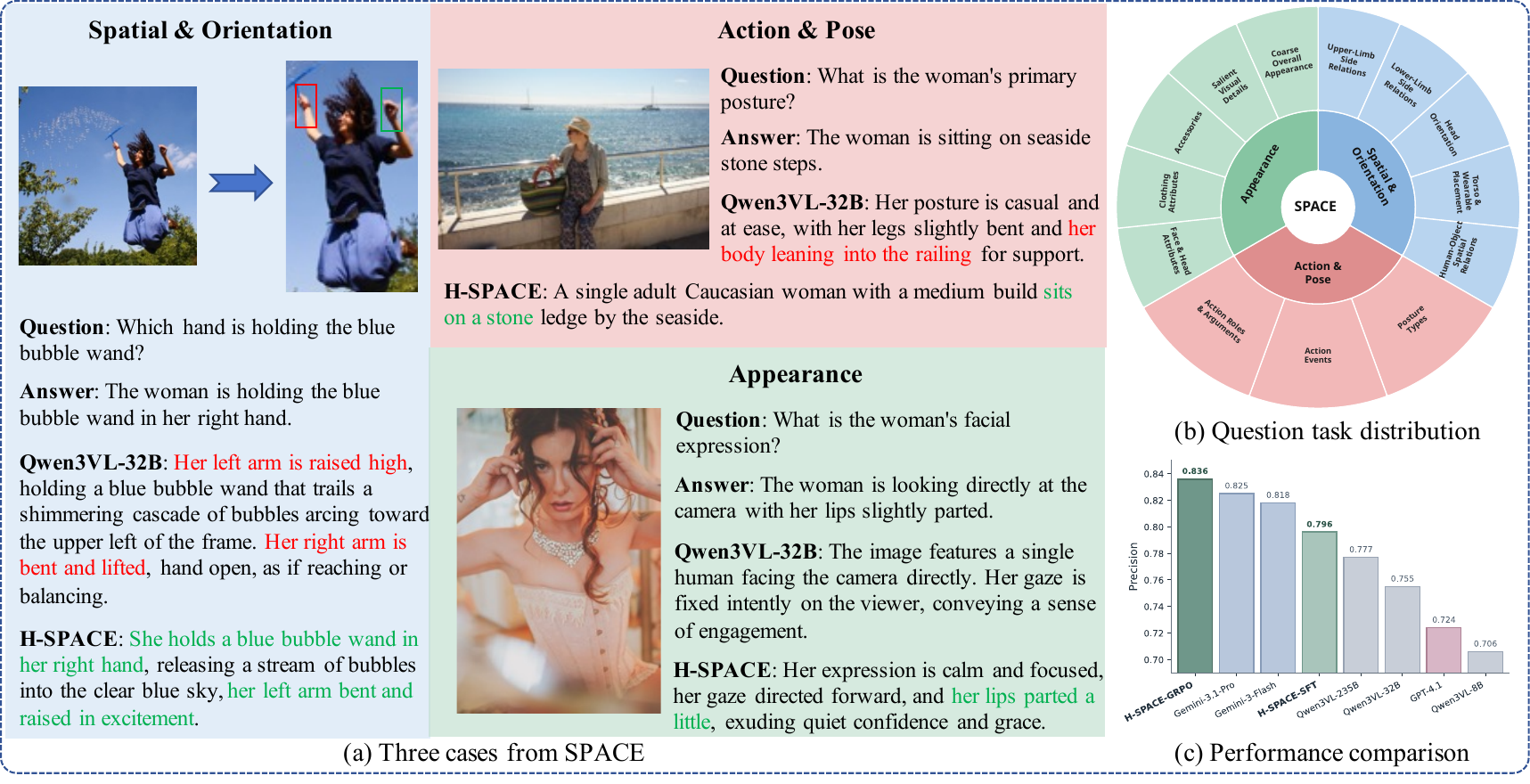}
    \caption{Overview of SPACE. (a) Three representative benchmark cases from the three categories: Spatial \& Orientation, Action \& Pose, and Appearance dimensions. Compared with Qwen3VL-32B, H-SPACE produces more accurate human-centric descriptions. (b) Question task distribution of SPACE, organized into three top-level dimensions and their fine-grained subcategories. (c) Performance comparison on SPACE using overall Precision. H-SPACE, especially after GRPO, achieves the best performance among representative Gemini, GPT, and Qwen baselines.}
    \Description{A three-panel overview of the SPACE benchmark. Panel (a) shows three example photographs of people, one per benchmark category, each paired with a caption from Qwen3VL-32B and a caption from H-SPACE; the Qwen3VL-32B captions contain spatial and body-part errors that the H-SPACE captions describe correctly. Panel (b) is a chart of how the benchmark questions are distributed over the three top-level dimensions and their fine-grained subcategories. Panel (c) is a bar chart of overall Precision on SPACE for representative Gemini, GPT and Qwen baselines alongside H-SPACE, in which H-SPACE after GRPO reaches the highest bar.}
    \label{fig:benchmark_casestudy}
\end{figure*}

%% file: section/0_intro.tex
\section{Introduction}
Multimodal large language models (MLLMs) have achieved substantial progress in general visual understanding and reasoning~\cite{comanici2025gemini,hurst2024gpt,bai2025qwen3,team2026kimi,zhang2025t5gemma,li2026qwen3,li2025gran,li2026semantic,hong2025glm,bai2024survey}. While recent advances have extended these capabilities toward detailed human analysis and fine-grained descriptions~\cite{wei2024large,liu2026humanmme,yang2022humancentric}, existing efforts still primarily emphasize broad human activities, coarse spatial relations, or overall caption richness. A central challenge in human-centric captioning remains underexamined: coordinate-frame conversion. Models observe an image from an extrinsic, camera-centered frame, whereas correct human descriptions often require an intrinsic, subject-centered frame. For example, accurately describing whether an accessory appears on a person's left or right side requires the model to convert from camera-centered observation to subject-centered description. This issue is important not only for faithful human-centered understanding, but also for constructing reliable human-centric caption corpora, whose quality directly affects the training of downstream multimodal and generative models~\cite{Wu2025QwenImageTR,Team2025ZImageAE,Gao2025Seedream3T}. However, existing caption benchmarks and human-centered evaluations do not explicitly test this conversion in generated descriptions, leaving a critical blind spot in current assessment.

To systematically study this issue, we introduce \textsc{SPACE} (Subject-centric Poses, Appearance, and Characteristics Evaluation), a comprehensive benchmark for human-centric image captioning. \textsc{SPACE} contains \textbf{2,981} images and \textbf{19,892} evaluation keypoints, organized into three categories: \textit{Spatial \& Orientation}, \textit{Action \& Pose}, and \textit{Appearance}. Among them, \textit{Spatial \& Orientation} is the core category and directly evaluates whether a model can translate camera-centered visual observations into accurate subject-centered descriptions. The other two categories, \textit{Action \& Pose} and \textit{Appearance}, provide complementary coverage of broader human-centric captioning ability, allowing us to distinguish failures in coordinate-frame conversion from more general captioning weaknesses. Through a comprehensive evaluation of mainstream MLLMs on \textsc{SPACE}, we reveal a clear vulnerability: despite strong performance on appearance and coarse actions, contemporary models consistently struggle with directional attribution, especially in distinguishing left and right body parts.

To address this limitation, we propose a scalable data construction and training framework for human-centric captioning. We first combine YOLO pose estimation~\cite{khanam2024yolov11} and SAM3-based~\cite{carion2025sam} part segmentation to derive structured body-part hints, including verified part-level bounding boxes, which provide explicit spatial evidence for annotation. Based on these hints, we decouple difficult human-centric caption construction into two stages: an initial captioning stage that produces coarse descriptions, and a rewriting stage in which a stronger proprietary MLLM revises the caption conditioned on the spatial hints to obtain subject-centered, spatially corrected supervision. We then distill this rewriting behavior into an open-source rewriter and use it for bootstrapped large-scale data expansion. Finally, we train a hint-free deployment model with supervised fine-tuning and Group Relative Policy Optimization (GRPO)~\cite{guo2025deepseek}. During preference optimization, a fine-grained rubric-based reward judge compares generated captions against rewritten references and assigns stronger penalties to spatially critical errors, enabling the model to better internalize subject-centered spatial constraints during generation.


Figure~\ref{fig:main} summarizes the overall pipeline of \textsc{SPACE} construction and model training.
\input{figures/figure1}

In summary, our main contributions are as follows:
\begin{itemize}[leftmargin=*]
    \item \textbf{Novel Benchmark and Insight:} We introduce \textsc{SPACE}, a comprehensive human-centric caption benchmark whose core \textit{Spatial \& Orientation} category explicitly targets coordinate-frame conversion (e.g., distinguishing a subject's left and right limbs), while \textit{Action \& Pose} and \textit{Appearance} provide complementary evaluation categories. Our extensive evaluation reveals that current mainstream models suffer from substantial deficiencies in accurately grounding directional body-part attributes.
    \item \textbf{Scalable Data Construction Pipeline:} We design an automated data synthesis workflow that harnesses expert vision-only models (YOLO and SAM3) for spatial prompt generation. Coupled with a two-stage caption-and-rewrite strategy and bootstrapped data expansion, this pipeline yields a large-scale training corpus with subject-centered spatial awareness.
    \item \textbf{Spatial-Aware Baseline Model:} Finally, we present a strong baseline model trained via SFT and standard GRPO with a rubric-based spatial judge reward. Experiments demonstrate that this model achieves state-of-the-art performance on \textsc{SPACE}, validating the efficacy of our automated data construction pipeline in instilling coordinate-frame awareness into MLLMs.

\end{itemize}

%% file: figures/figure1.tex
\begin{figure*}[t]
    \centering
    \IfFileExists{figures/main.pdf}{
        \includegraphics[width=0.86\textwidth]{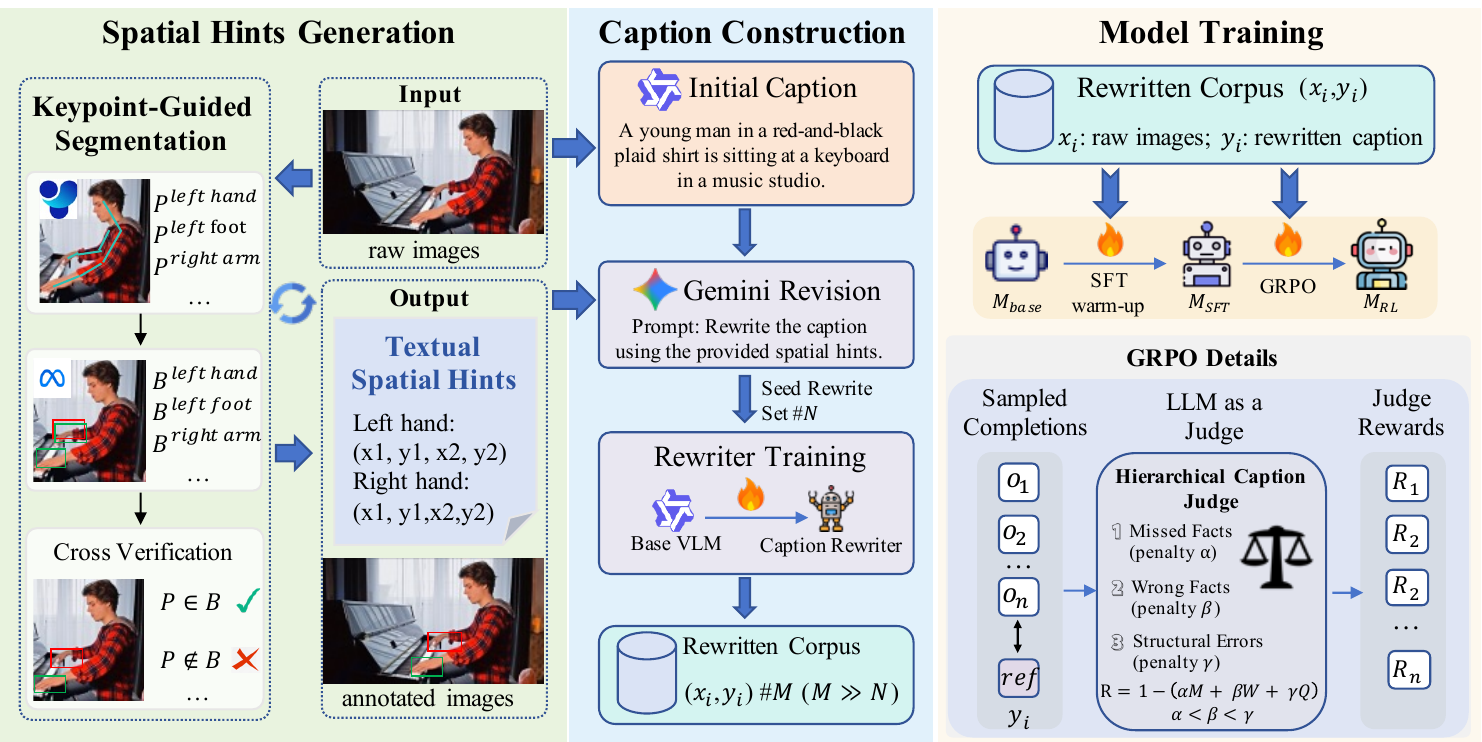}
    }{
        \fbox{\parbox[c][0.22\textheight][c]{0.95\textwidth}{\centering Main pipeline figure placeholder. Replace this box with \texttt{figures/main.pdf}.}}
    }
    \caption{Overview of our data construction and training pipeline. Left: YOLO pose estimation and SAM3-based part segmentation yield reliable body-part bounding boxes that serve as structured spatial hints. Middle: Gemini-3-Pro rewrites initial captions conditioned on these hints, and the rewritten data train an open-source rewriter for large-scale expansion. Right: the caption model is trained with supervised fine-tuning followed by GRPO, whose reward model scores generated captions against references.}
    \Description{A three-part pipeline figure. The left panel shows structured spatial hint generation by combining YOLO pose estimation and SAM3 segmentation to obtain reliable body-part boxes. The middle panel shows two-stage data construction with initial captioning, Gemini-3-Pro rewriting using caption and box hints, and open-source rewriter distillation for large-scale expansion. The right panel shows two-stage model training with supervised fine-tuning and GRPO, where a reward model scores generated captions against reference captions.}
    \label{fig:main}
\end{figure*}

%% file: section/1_related_work.tex
\section{Related Work}

\subsection{Human-Centric Benchmarking in MLLMs}

A large body of work has developed image and video captioning benchmarks to assess multimodal models from different perspectives, including descriptiveness, faithfulness, coverage, and human preference~\cite{liu2025capability,cheng-etal-2025-caparena,cafagna-etal-2023-hl,garg-etal-2024-imageinwords,chen2025avocado,chen2026diadem}. While these benchmarks have substantially improved caption evaluation, they are largely generic and are not tailored to human-centered understanding.

Recent human-centered benchmarks can be roughly divided into caption-oriented and understanding-oriented evaluations. Caption-oriented benchmarks remain relatively scarce, and existing efforts mainly emphasize human behaviors and human-object relations in generated descriptions~\cite{yang2022humancentric}. Understanding-oriented benchmarks instead assess human-centered multimodal understanding through question answering, grounding, or other discriminative formats~\cite{raza2025humanibench,liu2026humanmme}. Human-MME further broadens human-centered evaluation to diverse scenarios~\cite{liu2026humanmme}. However, these benchmarks still do not explicitly isolate coordinate-frame conversion in generated human descriptions, particularly subject-centered spatial errors such as left-right confusion and incorrect body-part binding. In contrast, \textsc{SPACE} focuses on human-centric evaluation with explicit attention to subject-centered spatial correctness in caption generation.

\subsection{Spatial Understanding and Frame of Reference in MLLMs}

A substantial body of literature has shown that, despite strong progress in general visual understanding tasks, current LVLMs still exhibit persistent weaknesses in spatial understanding and reasoning~\cite{rahmanzadehgervi2024vision,tong2024eyes,yue2024mmmu,liu2024mmbench,xie2024osworld,li2026unseen,li2025id,li2024bigger}. Existing efforts to improve spatial intelligence in multimodal models mainly follow two directions. One line injects explicit 3D or geometric signals, for example through geometry-aware encoders, reconstruction-oriented tokens, or explicit 3D intermediate supervision~\cite{wu2025spatial,fan2025vlm,chen2025think,ma2025spatialreasoner}. The other line improves spatial capability through targeted data curation and training paradigms, including spatial question-answer data, simulator-based generation, cognitive maps, staged supervised fine-tuning, reinforcement learning, and progressive curricula~\cite{wu2025spatial,ray2024sat,ouyang2025spacer,yin2025spatial,li2025spatialladder,yang2025cambrian}. Together, these studies have substantially advanced spatial grounding and spatial reasoning under generic visual settings.

Distinct from these general spatial modeling efforts, a specialized branch of research has emerged to investigate the psychological concept of \textit{Frame of Reference} (FoR) in MLLMs~\cite{Premsri_2025,Zhang2024DoVM,Zhang2025InSpireVM}. Cognitive science distinguishes between an extrinsic (camera-centered or viewer-centered) reference frame and an intrinsic (subject-centered or object-centered) reference frame. Specific benchmarks, such as COMFORT~\cite{Zhang2024DoVM} and FoREST~\cite{Premsri_2025}, explicitly reveal that while contemporary MLLMs perform reasonably well under extrinsic frames, they exhibit severe directional biases and substantial performance drops under intrinsic frames. Other recent efforts, such as InSpire~\cite{Zhang2025InSpireVM}, attempt to align Vision-Language-Action models with intrinsic object directionality to assist embodied agents. While these works highlight the FoR-related ``spatial blindspot'' of MLLMs, they mainly focus on VQA, classification, or embodied settings. In contrast, our work studies this coordinate-frame conversion problem in dense human-centric caption generation.

\subsection{Data Construction and Preference Optimization}

Recent work has increasingly explored automatic construction of richer caption supervision through multi-stage generation, question-guided enrichment, and rewrite-based refinement~\cite{zeng-etal-2025-enhancing-large,zhu2023chatgpt}. Related person-centered studies have further shown that MLLMs can be used to synthesize large-scale human descriptions, improve stylistic diversity, and reorganize person-centric annotations at scale ~\cite{tan2024harnessing,Jiang2025ModelingTO,qin2025human}. These efforts demonstrate the usefulness of synthetic supervision and rewriting pipelines, but they mainly aim to improve descriptiveness, diversity, or retrieval-oriented discriminability, rather than correcting subject-centered spatial errors with explicit geometric hints.

In parallel, recent post-training work has investigated reward-based optimization and preference alignment for captioning and LVLM generation, including metric-aware caption optimization, caption-specific preference learning, task-level preference alignment, and DPO-based mitigation of multimodal hallucination~\cite{takada-etal-2024-direct,huang-etal-2025-image,Yan2024TaskPO,Yang2025MitigatingHI}. More recently, rubric-based LLM and VLM evaluators, as well as rubric-oriented reward models, have highlighted the value of explicit scoring criteria for improving the interpretability and controllability of reward signals~\cite{Lee2024PrometheusVisionVM,Anugraha2025R3RR,Zhang2025ChasingTT,Jia2026OpenRS}. Benchmark studies on multimodal reward models further suggest that reliable visual preference signals remain task-sensitive and difficult to obtain~\cite{li2025vl,Zadeh2025LPOILP}. In contrast, our work focuses on human-centric caption generation under a subject-centered frame, and instantiates a task-specific rubric-based reward around spatial omissions, generic factual mistakes, and structurally critical orientation errors, while integrating structured spatial hints into both data construction and reward design.

%% file: section/2_benchmark.tex
\section{Proposed SPACE Benchmark}
\label{sec:benchmark}

\begin{table}[t]
\centering
\small
\caption{
Comparison between SPACE and prior human-centered spatial benchmarks.
Cat.: number of categories.
Imgs: number of images.
Subj.-centered: subject-centered orientation.
Non-subj. rel.: camera-centered spatial relations.
Body-part orient.: body-part-level orientation.
}
\setlength{\tabcolsep}{4pt}
\begin{tabular}{lccccc}
\toprule
\textbf{Benchmark} &
\textbf{Cat.} &
\textbf{Imgs} &
\makecell{\textbf{Subj.-}\\\textbf{centered}} &
\makecell{\textbf{Non-subj.}\\\textbf{rel.}} &
\makecell{\textbf{Body-part}\\\textbf{orient.}} \\
\midrule

HC-COCO-test & 1 & 2.5k & \textcolor{green}{\ding{51}} & \textcolor{red}{\ding{55}} & \textcolor{red}{\ding{55}} \\

HumaniBench & 7 & $\sim$1.5k & \textcolor{green}{\ding{51}} & \textcolor{red}{\ding{55}} & \textcolor{red}{\ding{55}} \\

Human-MME & 8 & 19,945 & \textcolor{green}{\ding{51}} & \textcolor{red}{\ding{55}} & \textcolor{green}{\ding{51}} \\

\midrule

\textbf{SPACE (Ours)} & 13 & 2,981
& \textcolor{green}{\ding{51}} & \textcolor{green}{\ding{51}} &\textcolor{green}{\ding{51}} \\

\bottomrule
\end{tabular}
\label{tab:benchmark_comparison}
\end{table}

\subsection{Motivation and Overview}

\zjh{While recent MLLMs demonstrate impressive capabilities in general object recognition, they consistently suffer from severe structural hallucinations when describing fine-grained human details. Among these, \textit{subject-centered spatial relations}—such as correctly distinguishing a subject's left and right limbs or accurately binding objects to specific body parts—represent a critical yet largely overlooked bottleneck. Although often overshadowed by overall descriptive accuracy in conventional evaluations, such structural inversions undermine the reliability of MLLMs in downstream applications such as human image/video generation, fine-grained action understanding, and embodied agent control. To systematically expose and quantify this vulnerability, we introduce \textit{SPACE} (\textbf{S}ubject-centric \textbf{P}oses, \textbf{A}ppearance, and \textbf{C}haracteristics \textbf{E}valuation).}

\textsc{SPACE} contains \textbf{2,981} images and \textbf{19,892} evaluation keypoints, organized into three top-level categories: \textbf{Spatial \& Orientation}, \textbf{Action \& Pose}, and \textbf{Appearance}. Among them, \textit{Spatial \& Orientation} is the core category and explicitly targets spatial relations involving human subjects. It covers both \textit{subject-centered orientation}, where the correct description must be made from the subject's perspective, and \textit{non-subject-centered spatial relations}, where the target relation is defined under a more generic or camera-centered frame. The other two categories, \textit{Action \& Pose} and \textit{Appearance}, provide complementary coverage of human-centric captioning and help distinguish whether model failures are specific to subject-centered spatial understanding or reflect broader weaknesses in human description generation. Figure~\ref{fig:benchmark_casestudy} provides a high-level overview of the benchmark composition and one representative benchmark instance. Model comparisons on \textsc{SPACE} are reported later in Section~\ref{sec:exp_main_results}.

Each sample is organized as an image-caption pair together with fine-grained evaluation keypoints and answers, which serve as the atomic evaluation units in \textsc{SPACE}. As shown later in Table~\ref{tab:benchmark_main_results}, current MLLMs exhibit the largest gap on the \textit{Spatial \& Orientation} category, especially on subject-centered cases, which motivates the dedicated spatially focused pipeline introduced in Section~\ref{sec:method_structured_prompt}.

\subsection{Benchmark Construction}


\zjh{Constructing a benchmark that specifically penalizes subject-cen\-tered spatial hallucinations requires a highly rigorous pipeline due to the serious hallucinations of current MLLMs. Automated metrics and LLM-generated captions frequently overlook or confuse egocentric left-right orientations. Therefore, we design a controlled construction process that relies heavily on manual verification to explicitly disentangle subject-centered semantics from generic visual descriptions.}

\noindent\textbf{1. Data Collection and Filtering:} We initiate the process by sampling raw images from the CapRL-5M~\cite{xing2025caprl} and LAION-400M~\cite{schuhmann2021laion} corpora. To ensure relevance, we apply a YOLO-based detector to filter the collected images, retaining only those containing human subjects. We preserve diverse human-centric scenes during this stage so that the benchmark covers varied appearances, poses, actions, and human-object configurations. Importantly, the images selected for \textsc{SPACE} are reserved for benchmark construction and are excluded from the training-data generation pipeline, ensuring that the benchmark and training corpus are strictly non-overlapping at the image level.

\noindent\textbf{2. Caption Generation and Manual Rectification:} For each filtered image, we prompt Gemini-3-Pro to generate an initial dense caption describing the human subjects and their activities. However, relying solely on automated generation often yields descriptions with subtle errors, particularly regarding spatial orientations and anatomical sides. Therefore, we perform meticulous manual corrections on these initial captions. Our annotators are instructed to rectify any descriptive inaccuracies and explicitly enforce a subject-centric coordinate system, ensuring that ambiguous body parts are accurately specified as ``left'' or ``right'' based on the subject's perspective. 

\noindent\textbf{3. Keypoint Generation for Evaluation:} Having established the rectified captions as ground-truth descriptions, we further derive fine-grained evaluation keypoints for each top-level category. For \textit{Spatial \& Orientation}, we predefine semantic points for both \textit{subject-centered orientation} and \textit{non-subject-centered spatial relations}. The former includes side-sensitive body-part cases such as hand/arm, leg/foot, head orientation, and torso-wearable bindings, while the latter covers human-object and local part-object relations under a more generic spatial frame. For \textit{Action \& Pose}, we instantiate keypoints over posture type, event verbs, and action arguments. Specifically, \textit{Appearance} is organized into \textit{Face and Head Cues}, \textit{Clothing and Style}, \textit{Accessories and Visible Attributes}, \textit{Salient Local Details}, and \textit{Overall Appearance Cues}. Conditioned on the verified captions, Gemini-3-Pro is then used to instantiate these keypoints and their corresponding ground-truth answers for each image.

This pipeline yields \textbf{2,981} image-caption pairs with \textbf{19,892} keypoint annotations for human-centric spatial and behavioral evaluation.

\subsection{Evaluation Protocol and Metrics}
Given a test image, the evaluated MLLM first generates a free-form caption, which is then scored against the predefined evaluation keypoints and their ground-truth answers by a judge model. The protocol follows the benchmark taxonomy introduced above. In particular, within \textit{Spatial \& Orientation}, Categories~1.1--1.4 correspond to subject-centered orientation cases that must be described from the subject's own frame, whereas Category~1.5 covers non-subject-centered spatial relations under a more generic frame. For each keypoint, the judge assigns one of three labels: \textit{correct}, meaning that the caption expresses the target semantic point correctly; \textit{wrong}, meaning that the caption mentions the target point but describes it incorrectly; and \textit{not mentioned}, meaning that the caption fails to cover the target point at all.

Let $\mathcal{K}$ denote a set of evaluation keypoints, which can correspond to the full benchmark or to a specific top-level category. We denote by $N_{\mathrm{c}}(\mathcal{K})$, $N_{\mathrm{w}}(\mathcal{K})$, and $N_{\mathrm{m}}(\mathcal{K})$ the numbers of keypoints judged as correct, wrong, and not mentioned, respectively. The total number of keypoints is therefore
\begin{equation}
|\mathcal{K}| = N_{\mathrm{c}}(\mathcal{K}) + N_{\mathrm{w}}(\mathcal{K}) + N_{\mathrm{m}}(\mathcal{K}).
\end{equation}
We report two complementary metrics. \textbf{Hit} measures coverage-quality over all evaluation points:
\begin{equation}
\mathrm{Hit}(\mathcal{K}) = \frac{N_{\mathrm{c}}(\mathcal{K})}{|\mathcal{K}|}.
\end{equation}
\textbf{Precision} evaluates correctness conditioned on the model having mentioned the relevant point:
\begin{equation}
\mathrm{Precision}(\mathcal{K}) =
\frac{N_{\mathrm{c}}(\mathcal{K})}
{N_{\mathrm{c}}(\mathcal{K}) + N_{\mathrm{w}}(\mathcal{K})}.
\end{equation}
Intuitively, Hit penalizes both omission and error, while Precision isolates whether the model is correct once it attempts to describe the corresponding semantic point. We report these two metrics for the overall benchmark and for each top-level category separately.

%% file: section/3_method.tex
\section{The Proposed Method}

\begin{table*}[!htbp]
\centering
\small
\setlength{\tabcolsep}{6pt}
\renewcommand{\arraystretch}{1.12}
\caption{Main results on \textsc{SPACE}. We report \textit{Precision} and \textit{Hit} for the overall benchmark and the three top-level dimensions. Models are grouped into closed-source and open-source MLLMs; within each group, the best value per column is bolded. Parameter counts are given when publicly known; ``--'' denotes an undisclosed model size.}
\label{tab:benchmark_main_results}
\begin{tabular}{@{}l@{\hspace{8pt}}c@{\hspace{18pt}}c@{\hspace{3pt}}c@{\hspace{18pt}}c@{\hspace{3pt}}c@{\hspace{18pt}}c@{\hspace{3pt}}c@{\hspace{18pt}}c@{\hspace{3pt}}c@{}}
\toprule
\textbf{Model} & \textbf{Params} & \multicolumn{2}{c}{\textbf{Overall}} & \multicolumn{2}{c}{\textbf{Spatial \& Ori.}} & \multicolumn{2}{c}{\textbf{Action \& Pose}} & \multicolumn{2}{c}{\textbf{Appearance}} \\
\cmidrule(lr){3-4}\cmidrule(lr){5-6}\cmidrule(lr){7-8}\cmidrule(lr){9-10}
 &  & \textbf{Prec.} & \textbf{Hit} & \textbf{Prec.} & \textbf{Hit} & \textbf{Prec.} & \textbf{Hit} & \textbf{Prec.} & \textbf{Hit} \\
\midrule
\multicolumn{10}{l}{\textbf{Closed-source Models}} \\
\textit{GPT-4.1}                                      & --   & 72.4 & 44.5 & 67.9 & 47.7 & 72.0 & 56.4 & \textbf{84.4} & 28.5 \\
\textit{Gemini-3-Flash}~\cite{comanici2025gemini}     & --   & 81.8 & 62.1 & 80.1 & 67.0 & \textbf{83.5} & 70.3 & 82.1 & 47.8 \\
\textit{Gemini-3.1-Pro}~\cite{comanici2025gemini}     & --   & \textbf{82.5} & \textbf{67.6} & \textbf{81.1} & \textbf{74.5} & 83.1 & \textbf{75.6} & 84.0 & \textbf{51.3} \\
\addlinespace[2pt]
\midrule
\multicolumn{10}{l}{\textbf{Open-source Models}} \\
\textit{MiMo-VL-7B-SFT}~\cite{Yue2025MiMoVLTR}         & 7B   & 70.2 & 40.8 & 68.4 & 42.7 & 68.5 & 49.6 & 76.6 & 29.3 \\
\textit{MiniCPM-V-4.5}~\cite{yu2025minicpm}           & 9B   & 67.5 & 43.7 & 63.4 & 45.4 & 68.1 & 51.8 & 73.9 & 33.5 \\
\textit{Ovis2.5-9B}~\cite{lu2025ovis2}                & 9B   & 72.6 & 47.1 & 68.2 & 47.5 & 72.9 & 56.0 & 79.6 & 37.6 \\
\textit{InternVL3.5-8B}~\cite{wang2025internvl3}      & 8B   & 64.3 & 35.5 & 60.5 & 35.1 & 63.7 & 43.9 & 72.1 & 27.3 \\
\textit{InternVL3.5-14B}~\cite{wang2025internvl3}     & 14B  & 63.9 & 32.4 & 60.9 & 32.1 & 64.3 & 42.7 & 68.9 & 22.0 \\
\textit{Qwen3-VL-8B-Instruct}~\cite{bai2025qwen3}    & 8B   & 70.6 & 56.7 & 66.6 & 59.1 & 72.8 & 63.2 & 73.9 & 47.0 \\
\textit{Qwen3-VL-8B-Thinking}~\cite{bai2025qwen3}     & 8B   & 73.3 & 51.8 & 70.4 & 56.7 & 72.9 & 60.5 & 79.6 & 37.3 \\
\textit{Qwen3-VL-32B-Instruct}~\cite{bai2025qwen3}    & 32B  & 75.5 & 61.3 & 72.1 & 65.3 & 76.4 & 68.2 & 79.8 & 49.6 \\
\textit{Qwen3-VL-235B-A22B-Instruct}~\cite{bai2025qwen3} & 235B & 77.7 & 60.7 & 74.9 & 65.0 & 78.2 & 68.2 & 82.1 & 47.8 \\
\addlinespace[2pt]
\midrule
\textbf{H-SPACE-GRPO}                                  & 8B   & \textbf{83.6} & \textbf{67.9} & \textbf{82.4} & \textbf{70.4} & \textbf{84.0} & \textbf{68.9} & \textbf{84.6} & \textbf{63.9} \\
\bottomrule
\end{tabular}
\end{table*}

Figure~\ref{fig:main} provides an overview of our method. As shown by the benchmark results in Table~\ref{tab:benchmark_main_results} and Figure~\ref{fig:category11_line}, current MLLMs exhibit the largest weakness on the \textit{Spatial \& Orientation} category, especially on subject-centered cases. Motivated by this difficulty, we design a dedicated pipeline that first constructs reliable spatial hints, then produces spatially corrected training captions, and finally optimizes a deployment model with rubric-based preference learning.
\subsection{Structured Spatial Hint Generation}
\label{sec:method_structured_prompt}

This stage converts each image into conservative structured spatial hints for later caption correction. For each detected person, we first use YOLO11x-Pose to estimate a human skeleton, then use SAM3 to localize body-part candidates inside the person crop, and finally apply a side-aware geometric verification rule to remove unreliable left-right assignments. The output of this stage is a hint-annotated image together with a textual hint list.

\noindent\textbf{1. Human Pose Estimation (YOLO):} 
Given an input human-centric image $\mathcal{I}$, we apply a pre-trained YOLO-Pose model to detect human subjects and annotate their skeletal structure. Let $f_{\mathrm{YOLO}}$ denote the pose estimation function. For each detected person, YOLO-Pose returns a person box together with a set of $N$ anatomical joints $\mathcal{J} = \{ j_i = (x_i, y_i, c_i) \}_{i=1}^N$, where $(x_i, y_i)$ represents the 2D coordinate of the $i$-th joint (e.g., left wrist) and $c_i$ is the confidence score. These outputs tell us where the subject is and provide coarse anatomical anchors for later part-level localization.

\noindent\textbf{2. Part Candidate Localization (SAM3):}
For each detected person, we crop the image to the YOLO person box and run SAM3 ($f_{\mathrm{SAM}}$) only inside this localized region. Using body-part-specific textual prompts, SAM3 produces candidate masks and bounding boxes for the queried parts. If SAM3 returns multiple candidates for the same part, we keep the candidate whose box center is closest to the corresponding YOLO keypoint. Therefore, YOLO keypoints are not fed to SAM3 as prompts; instead, they act as anatomical anchors for candidate selection.

\noindent\textbf{3. Side-Aware Verification and Hint Construction:}
The last step is to verify whether the SAM3 candidates are geometrically consistent with the YOLO skeleton, especially for bilateral parts such as hands, feet, and shoulders. Let $r_{\ell}$ and $r_{r}$ denote the SAM3 candidates initially associated with the left and right side of a bilateral part, and let $p_{\ell}$ and $p_{r}$ denote the corresponding YOLO keypoints when available. For any candidate $r$, we denote the center of its bounding box by $\mathbf{c}(r)$ and measure geometric compatibility by the Euclidean distance $d(\mathbf{c}(r), p)$.

The rule is intentionally conservative. When both side keypoints are available, we keep a candidate on its nominal side only if it is closer to the same-side keypoint than to the opposite-side one. If both candidates collapse toward the same side, we keep only the better-supported candidate for that side and mark the opposite side as unresolved. If the two candidates cross over, we discard both rather than swap them. When only one side keypoint is visible, we preserve the nominal candidate only when it is the closer one; otherwise both sides remain unresolved. If the corresponding keypoint is unavailable, the part is treated as unresolved and omitted from the final hint set. This rule follows a simple principle: a missing spatial hint is preferable to a confidently wrong left-right hint.

Let $\mathcal{L}$ denote the predefined set of queried body-part labels, and let $b_L$ denote the final retained bounding box for part label $L \in \mathcal{L}$ after verification, where $b_L=\varnothing$ indicates that the part is unresolved and therefore discarded. We then define the valid body-part set as
\begin{equation}
    \mathcal{L}_{\mathrm{valid}} = \left\{ L \in \mathcal{L} \mid b_L \neq \varnothing \right\},
\end{equation}
and collect the verified boxes as
\begin{equation}
    \mathcal{B}_{\mathrm{verified}} = \left\{ b_L \mid L \in \mathcal{L}_{\mathrm{valid}} \right\}.
\end{equation}

For each retained body part, we draw the verified box directly onto the raw image to obtain a hint-annotated image $\mathcal{I}_{\mathrm{draw}}$. To make subject-centered spatial cues easier to parse, we use different colors to distinguish bounding boxes associated with different subject-centered side assignments (e.g., left versus right) and other spatial roles. In parallel, we serialize the corresponding part labels and coordinates as a textual hint list
\begin{equation}
    \mathcal{C} = \left\{ (L, b_L) \mid L \in \mathcal{L}_{\mathrm{valid}} \right\}.
\end{equation}
Together, $(\mathcal{I}_{\mathrm{draw}}, \mathcal{C})$ constitute the structured spatial hints used in the subsequent data generation and training stages.

\subsection{Two-Stage Caption Generation and Rewriting Pipeline}
\label{sec:method_data_construction}

To construct high-quality, human-centric image-caption data that accurately reflects the subject-centered coordinate system, we design a dedicated two-stage data generation workflow. Generating a fully correct human-centric caption in a single pass is intrinsically difficult because it requires both broad semantic coverage and precise subject-centered spatial reasoning. We therefore decompose the problem into an initial description phase followed by a structured refinement phase. This decomposition separates broad human description from spatially critical correction, which simplifies data construction while preserving semantic coverage.

\noindent\textbf{1. Initial Caption Generation:} 
To establish the data source for our rewriting pipeline, we first collect raw images from CapRL-5M and LAION-400M. We then apply a YOLO object detector for preliminary filtering to isolate images containing human subjects. The training corpus is constructed from a subset that is strictly disjoint from the images used in \textsc{SPACE}, so that no benchmark image is reused during model training. For each retained human-centric image $\mathcal{I}$, we prompt Qwen3-VL-235B-A22B to generate a detailed human-centric caption, which serves as the initial caption for the subsequent rewriting stage.

\noindent\textbf{2. Structured Prompt-Conditioned Rewriting:} 
To rectify the spatial inaccuracies in the initial caption, we employ a proprietary MLLM, Gemini-3-Pro, as a rewriter. We provide the model with the hint-annotated image $\mathcal{I}_{\mathrm{draw}}$, the textual hint list $\mathcal{C}$, and the initial caption. Conditioned on these structured spatial hints, the rewriter revises the caption so that body parts are more explicitly grounded to their corresponding visual regions and directional expressions are more consistent with the subject-centered frame.

\noindent\textbf{3. Distillation and Bootstrapped Data Expansion:}
Relying entirely on a proprietary model for large-scale data annotation is cost-prohibitive. To scale up the corpus, we use the rewritten captions to fine-tune an open-source MLLM, distilling the rewriting behavior into a lower-cost model. Once trained, this model can use the same structured spatial hints to produce refined captions at scale. We then apply it for bootstrapped data expansion over a broader set of human-centric images, yielding a larger rewritten corpus for final training. For each rewriting instance, $x_i$ denotes the raw-image captioning input, $x_i^{\mathrm{hint}}$ denotes the auxiliary hint-augmented input used only by the rewriter, and $y_i^\star$ denotes the rewritten target caption. The resulting deployment-training corpus therefore retains the rewritten pairs $(x_i, y_i^\star)$ rather than the intermediate hint-conditioned inputs.

\subsection{Supervised Fine-Tuning and Preference Optimization}
\label{sec:method_training}

Given the rewritten tuples $(x_i, y_i^\star)$, we train a unified multimodal deployment model $\pi_\theta$ with supervised fine-tuning followed by GRPO.

\noindent\textbf{1. Supervised Fine-Tuning (SFT):}
We first optimize $\pi_\theta$ on the rewritten corpus using teacher forcing:
\begin{equation}
    \mathcal{L}_{\mathrm{SFT}} = - \sum_{t=1}^{T} \log \pi_\theta(y_{i,t}^\star \mid x_i, y_{i,<t}^\star).
\end{equation}
We then freeze the resulting model as the reference policy $\pi_{\mathrm{ref}}$ for the subsequent preference optimization stage.

\noindent\textbf{2. Hierarchical Rubric-Based Judge Reward:}
Our main task-specific design lies in the reward. Given the rewritten target $y_i^\star$, a judge model extracts a set of expected facts $E_i$ and compares them with a sampled caption, yielding three disjoint error sets: missed facts $M_i$, generic wrong facts $W_i$, and critical structural errors $Q_i$~\cite{Lee2024PrometheusVisionVM,Anugraha2025R3RR,Zhang2025ChasingTT,Jia2026OpenRS}. The last category captures the failures that matter most in our setting, especially left-right anatomical reversals and incorrect subject-object binding. We therefore assign hierarchical penalties $\alpha < \beta \ll \gamma$ and define the reward as
\begin{equation}
R_i^{\mathrm{judge}}
=
\operatorname{clip}\!\left(
1-\frac{\alpha |M_i|+\beta |W_i|+\gamma |Q_i|}{\max(1, |E_i|)},
R_{\min}, 1
\right).
\end{equation}
In GRPO, this score is used as the reward $r_i^{(g)}$ for each sampled completion.

\noindent\textbf{3. Group Relative Policy Optimization (GRPO):}
We use standard GRPO to optimize $\pi_\theta$ against the frozen reference policy $\pi_{\mathrm{ref}}$. The objective is
\begin{equation}
\begin{aligned}
\ell_{i,g,t}^{\mathrm{GRPO}}
&=
- \min\!\Big(\rho_{i,g,t} A_i^{(g)}, \operatorname{clip}(\rho_{i,g,t}, 1-\epsilon, 1+\epsilon) A_i^{(g)}\Big)
\\
&\qquad\qquad
+ \lambda_{\mathrm{KL}} \, \mathcal{D}_{\mathrm{KL}}(\pi_\theta \,\|\, \pi_{\mathrm{ref}})
\;.
\end{aligned}
\end{equation}
where
\[
\rho_{i,g,t}
=
\frac{\pi_\theta(\hat{y}_{i,t}^{(g)} \mid x_i)}
{\pi_{\mathrm{old}}(\hat{y}_{i,t}^{(g)} \mid x_i)}.
\]
Here, $A_i^{(g)}$ denotes the group-wise relative advantage. At inference time, we discard all training-time hints and deploy the model directly on the raw-image input $x_i$.

%% file: section/4_exp.tex
\section{Experiments}

\begin{figure*}[!ht]
    \centering
    \includegraphics[width=0.86\textwidth]{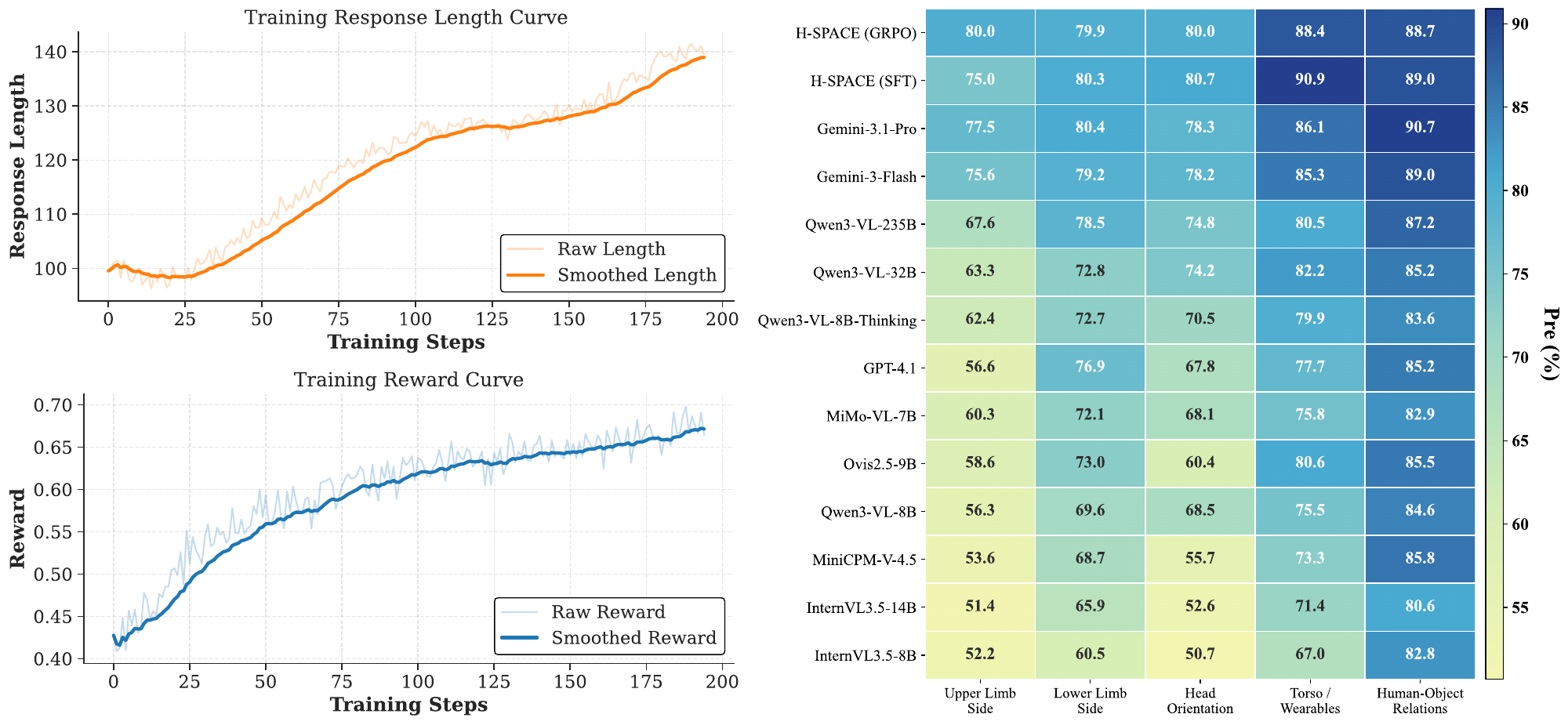}
\caption{Left: training dynamics of our model during post-training, including the response-length curve (top) and the reward curve (bottom), both of which show a steady upward trend over training steps. Right: fine-grained precision heatmap on Category~1 (\textit{Spatial \& Orientation}) of \textsc{SPACE}.}
\Description{On the left, two line charts of post-training dynamics plotted against training steps: the upper chart shows response length rising steadily, and the lower chart shows reward rising steadily; each chart plots a raw curve and a smoothed curve. On the right, a heatmap of Precision on the fine-grained subcategories of Category~1, with one row per evaluated model and one column per subcategory; darker cells denote higher precision, and the H-SPACE-GRPO row is the darkest overall.}
\label{fig:category11_line}
\end{figure*}

\subsection{Datasets and Implementation Details}
We evaluate all models on \textsc{SPACE} under the protocol described in Section~\ref{sec:benchmark}, using \textit{Qwen3-VL-235B-A22B-Instruct} as the judge model. Unless otherwise specified, caption generation uses temperature $0$.

For data construction, we use \textit{YOLO11x-Pose} and \textit{SAM3} for structured hint extraction. We employ \textit{Gemini-3-Pro} to annotate approximately \textbf{60K} human-centric samples, distill the rewriting behavior into an open-source rewriter based on \textit{Qwen3-VL-32B}, and further expand the rewritten corpus to approximately \textbf{237K} samples. The final deployment model is built on \textit{Qwen3-VL-8B-Instruct}.

All training stages use \textit{AdamW} with a cosine learning-rate schedule, warmup ratio \textbf{0.03}, and \textbf{1} epoch. For supervised fine-tuning, we use batch size \textbf{128}, learning rate \textbf{$1\times10^{-5}$}, and \textbf{187K} rewritten samples. For GRPO, we use learning rate \textbf{$1\times10^{-6}$}, KL coefficient $\lambda_{\mathrm{KL}}=\textbf{0.03}$, \textbf{50K} training samples, \textbf{8} sampled completions per prompt, and \textit{Qwen3-VL-32B-Instruct} as the rubric-based reward judge. The reward coefficients are $\alpha_{\mathrm{missed}}=\textbf{1.0}$, $\beta_{\mathrm{wrong}}=\textbf{1.5}$, $\gamma_{\mathrm{critical}}=\textbf{2.5}$, with $R_{\min}=\textbf{-1.0}$.

\subsection{Main Results}
\label{sec:exp_main_results}

Table~\ref{tab:benchmark_main_results} presents the main results on \textsc{SPACE}, comparing representative open-source and closed-source MLLMs under the same prompt and judge-based protocol, with \textit{Precision} and \textit{Hit} reported over the full benchmark and the three top-level dimensions. H-SPACE-GRPO achieves the highest overall performance among the compared models, reaching \textbf{83.6} Precision and \textbf{67.9} Hit. Relative to the base Qwen3-VL-8B-Instruct, it improves overall Precision and Hit by \textbf{13.0} and \textbf{11.2} points, respectively. It also performs better than substantially larger open-source models such as Qwen3-VL-32B-Instruct and Qwen3-VL-235B-A22B-Instruct, suggesting that task-specific data construction and post-training are effective for human-centric captioning under a subject-centered frame.

H-SPACE-GRPO also compares favorably with strong closed-source systems. In particular, it slightly exceeds Gemini-3.1-Pro on the overall benchmark (\textbf{83.6} vs.\ 82.5 Precision, \textbf{67.9} vs.\ 67.6 Hit), and attains the highest Precision in all three top-level dimensions. Notably, H-SPACE-GRPO achieves the best Precision on \textit{Spatial \& Orientation} (\textbf{82.4}), which is consistent with the goal of reducing structurally critical subject-centered errors. At the same time, Gemini-3.1-Pro remains stronger in Hit on \textit{Spatial \& Orientation} and \textit{Action \& Pose}, indicating that coverage of relevant semantic points is still not fully solved.

\subsection{Fine-Grained Spatial Analysis}
\label{sec:exp_finegrained}

To better understand where models fail within the core \textit{Spatial \& Orientation} dimension, we examine a fine-grained breakdown of its representative subcategories, which reveals whether the observed weakness is broadly distributed or concentrated on a smaller set of subject-centered cases. Figure~\ref{fig:category11_line} visualizes the relative performance of different models on these categories.

Across models, Category~1.1 emerges as one of the most challenging cases, whereas Category~1.5 is generally handled much better. Compared with Category~1.5, Category~1.1 requires explicit subject-centered left-right discrimination of the arms and hands, making it particularly sensitive to failures in intrinsic frame conversion. Several open-source models achieve only slightly above 50\% Precision on Category~1.1 while showing substantially stronger performance on Category~1.5. This contrast suggests that subject-centered spatial grounding, rather than generic human description, remains a major bottleneck in human-centric captioning.

This trend holds across model families, so the difficulty is not tied to a specific architecture or model scale. Current MLLMs are therefore not uniformly weak on human-centric captioning; their failures concentrate on the cases that require reliable intrinsic frame conversion and precise body-part grounding, which reinforces the main finding of \textsc{SPACE}.

\subsection{Controlled Rewriting Analysis}
\label{sec:exp_rewriting}
We further analyze the effect of caption rewriting in a controlled setting. We fix the source captions to those produced by a Qwen3-VL baseline on \textsc{SPACE} and only vary the rewriting stage.

We consider three rewriting variants. First, we use \textit{Gemini-3.1-Pro} as a rewriter without structured hints, conditioning only on the original image and the source caption. Second, we provide \textit{Gemini-3.1-Pro} with the full hint-conditioned input, including the hint-annotated image and the bbox-derived textual hints, to probe the direct effect of adding structured spatial guidance. Third, we evaluate our trained rewriter under the same hint-conditioned input, allowing us to assess both the effectiveness of the learned rewriter itself and whether hint-guided rewriting behavior can be realized in a lower-cost open-source model.

Because the source captions are fixed across settings, the observed differences can be attributed to the rewriting stage itself rather than to variation in the base captioner. Concretely, on Category~1, \textit{Gemini-3.1-Pro} without hints, \textit{Gemini-3.1-Pro} with hints, and our trained rewriter obtain \textbf{73.3}, \textbf{80.7}, and \textbf{80.0} Precision, respectively. These results support two observations. First, structured hints are informative: supplying them to Gemini-3.1-Pro improves Category~1 Precision by \textbf{7.4} points, indicating that the bbox-derived spatial guidance carries subject-centered information that even a strong rewriter cannot reliably recover from the image and the source caption alone. Second, our trained rewriter is effective: despite being an open-source distilled model, it reaches \textbf{80.0} Precision under the same hint-conditioned input, essentially matching the proprietary rewriter at substantially lower cost. Overall, subject-centered spatial errors are not only a challenging evaluation target, but also a correctable one given effective rewriting and hint utilization.

%% file: section/5_conclusion.tex
\section{Conclusion}
We study subject-centered spatial understanding in human-centric image captioning. We introduce \textsc{SPACE}, a benchmark that evaluates whether models can convert camera-centered observations into correct subject-centered descriptions, on which current MLLMs still struggle despite strong performance on broader visual tasks. To address this gap, we propose a scalable data construction and training pipeline combining structured spatial hints, caption rewriting, and rubric-based preference optimization. The resulting model substantially improves caption quality on \textsc{SPACE}, particularly in reducing subject-centered spatial errors, and remains competitive with strong closed-source systems.